\documentclass[conference]{IEEEtran}
\IEEEoverridecommandlockouts
\usepackage{cite}
\usepackage{amsmath,amssymb,amsfonts}
\usepackage{algorithmic}
\usepackage{graphicx}
\usepackage{textcomp}
\usepackage{xcolor}
\usepackage{commath}
\usepackage{siunitx}
\usepackage{caption}
\usepackage{subcaption}
\usepackage[belowskip=-5pt,aboveskip=9pt]{caption}
\usepackage{romannum}
\usepackage{multirow}
\usepackage{tabularx}
\usepackage{booktabs}
\usepackage{url}

\def\BibTeX{{\rm B\kern-.05em{\sc i\kern-.025em b}\kern-.08em
    T\kern-.1667em\lower.7ex\hbox{E}\kern-.125emX}}
\begin{document}

\title{Optimizing multi-market participation of battery and electrolyser systems based on field performance}

\author{\IEEEauthorblockN{Chunyang Zhao, Stoyan Trenchev, Shi You, Chresten Træholt}
\IEEEauthorblockA{\textit{Department of Wind and Energy Systems} \\
\textit{Technical University of Denmark}\\
Kongens Lyngby, Denmark \\
chuzh@dtu.dk, stoyan.trenchev@mailbox.org, shyo@dtu.dk, ctra@dtu.dk}
}

\maketitle

\begin{abstract}
The increasing share of renewable energy in power systems creates a need for fast-response and flexible resources to maintain system stability. With the expansion of electricity markets and ancillary service products, opportunities arise to stack revenues across multiple services. Long-term Power-to-X (PTX) electrolysers and short-term battery energy storage systems (BESS) are prevalent flexible resources, yet most studies neglect real hardware behavior, such as ramp limits, efficiency, setpoint-tracking accuracy, etc. This work presents experimental and modeling results for a 55 kW/79 kWh BESS and an electrolyzer comprising three 2.4 kW units. Key characteristics are identified through measurements and embedded into a price-driven optimization framework for participation in the Danish electricity and ancillary service markets, utilizing real market data from 2022 to 2025. The optimized daily profits for multi-market participation are 1,749.27 DKK and 289.46 DKK for BESS and the electrolyser, respectively. With the demonstrated business cases for BESS and PTX systems, this work addresses the importance of incorporating experimental performance across multiple markets and years.
\end{abstract}

\begin{IEEEkeywords}
Battery energy storage system (BESS), Power-to-X (PTX), Energy market, Ancillary service.
\end{IEEEkeywords}

\section{Introduction}\label{sec:introduction}

The global energy transition is characterized by rapid electrification and an accelerating deployment of variable renewable energy sources such as wind and solar photovoltaic (PV) \cite{ShivaramaKrishna2015ASystems}.  While large‐scale plants continue to dominate new capacity additions, the penetration of distributed energy resources (DERs) at the grid edge, such as battery energy storage systems (BESS), heat pumps, and electrolysers, is increasingly recognized as essential for balancing supply and demand, mitigating grid congestion, and providing system services \cite{Zhao2023Grid-connectedIntegration}. Aggregated DERs can offer fast, localized flexibility and complement conventional reserves traditionally provided by synchronous generators, while Power-to-X assets can produce hydrogen to contribute to broader decarbonization goals \cite{Jin2025ExperimentsReview}.

In parallel, the Nordic power system is introducing new market products that reward fast response and bidirectional flexibility. Services such as Fast Frequency Reserve (FFR), Frequency‐Controlled Disturbance Reserve (FCRD), and Frequency‐Controlled Normal Operation Reserve (FCRN) are now open to various flexible energy resources, provided that they meet activation‐time, duration, and accuracy requirements \cite{en.energinet.dkMarketRequirements}. These developments create new business opportunities for small, responsive assets that can participate simultaneously in multiple markets, a strategy often referred to as multi‐market optimization or revenue stacking \cite{Hjalmarsson2023ServiceReview}.

However, most existing studies rely on idealized or specification data and neglect the physical limitations of real hardware \cite{Tofighi-Milani2025Electrolysers:Responses}. Specifically, laboratory and field assets exhibit constrained voltage windows, temperature-dependent efficiencies, limited setpoint tracking accuracy, and nonlinear start-up or ramp dynamics that can significantly influence market eligibility and profitability \cite{Zhao2024Lab-fieldSystem}. For example, the battery state of charge (SOC) and state of energy (SOE) can exhibit a significant divergence in large-scale BESS, and the start-up behavior of DER assets cannot be represented by simple electrical or physical models.
Accurately quantifying these limitations and embedding them into dispatch algorithms is essential for realistic performance assessment.

This paper addresses this gap by experimentally characterizing and modeling two real assets installed at the Technical University of Denmark: a modular Xolta BESS rack and a modular anion exchange membrane (AEM) electrolyzer. The contributions are threefold:
\begin{enumerate}
\item Experimental quantification of each asset’s operational constraints relevant to market participation, including usable energy range, round‐trip efficiency, setpoint accuracy, preheat behavior, ramp rates, etc.
\item Development of a data‐driven framework that optimizes energy and reserve bids across the Danish day‐ahead and ancillary‐service markets using historical data.
\item Optimized dispatch to evaluate revenue potential based on the real-world hardware operation constraints for multiple markets from days to years.
\end{enumerate}
In this work, we demonstrate that laboratory-scale energy assets can be optimally dispatched across multiple electricity markets when their physical constraints are explicitly modeled. After reviewing the state-of-the-art in Section~\ref{sec:introduction},  comprehensive experimental characterizations of both the BESS and the electrolyzer are presented, capturing their short-term dynamic behavior as well as long-term performance limitations in Section~\ref{sec:hardware}. In Section~\ref{sec:modeling}, the multi-market optimization framework and operational objectives are detailed. In Section~\ref{sec:results}, simulation results are illustrated, including daily optimization and multi-year validation. The paper concludes with discussing key findings and broader implications in Section~\ref{sec:Discussion} \& \ref{sec:conclusion}.

\section{Experiments and Asset Characterization}\label{sec:hardware}
\begin{table}[tb]
\centering
\caption{Key parameters of the assets.}
\label{tab:assetparams}
\setlength{\tabcolsep}{4pt} 
\begin{tabular}{lcc}
\toprule
\textbf{Parameter} & \textbf{BESS} & \textbf{Electrolyzer} \\ \midrule
Power & \SI{55}{kW} (\SI{30}{kW} used) & 3 $\times$ \SI{2.4}{kW} \\
Energy & \SI{79}{kWh} (\SI{52}{kWh} used) & -- \\
Efficiency & 90--98\% RTE & 62--70\% HHV \\
Response time (up/down) & \SI{2}{s} / \SI{2}{s} & \SI{21}{s} / \SI{1.6}{s} \\
Preheat / standby power & -- & \SI{0.2}{kW} per module \\
SOC / power limits & 12--95\% SOC & 60--100\% $P_{\text{nom}}$ \\
State of health (2025) & 90\% & 100\% \\
\bottomrule
\end{tabular}
\end{table}

Our PowerLabDK platform hosts a wide range of industrial hardware for testing renewable energy and storage technologies under realistic conditions, including off-grid, direct connection to the local Danish grid, and a controllable grid via a power amplifier. Details can be found in \cite{Zhao2024Lab-fieldSystem}. In this study, two assets were involved: a modular BESS for the grid application and a modular AEM electrolyzer for hydrogen production. Both devices can be scaled up for the MW-level grid-connected system and equipped with high-fidelity monitoring and remote control, enabling the combination of experimental measurements with data-driven market modeling. Table \ref{tab:assetparams} summarizes the experimental parameters used in the subsequent market optimization and back-testing.
\begin{figure}[b]
    \centering
    \includegraphics[width=1\linewidth]{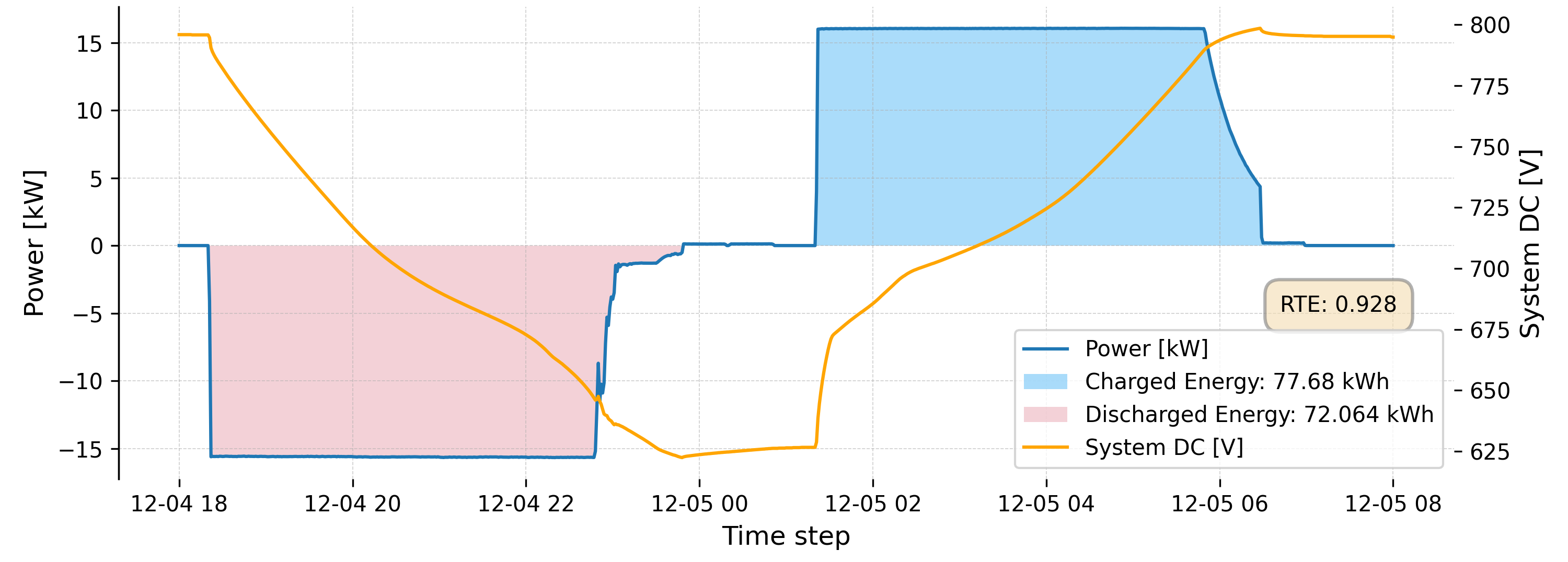}
    \caption{BESS round-trip test at 15 kW}
    \label{fig:roundtrip}
\end{figure}
\begin{figure}[b]
    \centering
    \includegraphics[width=1\linewidth]{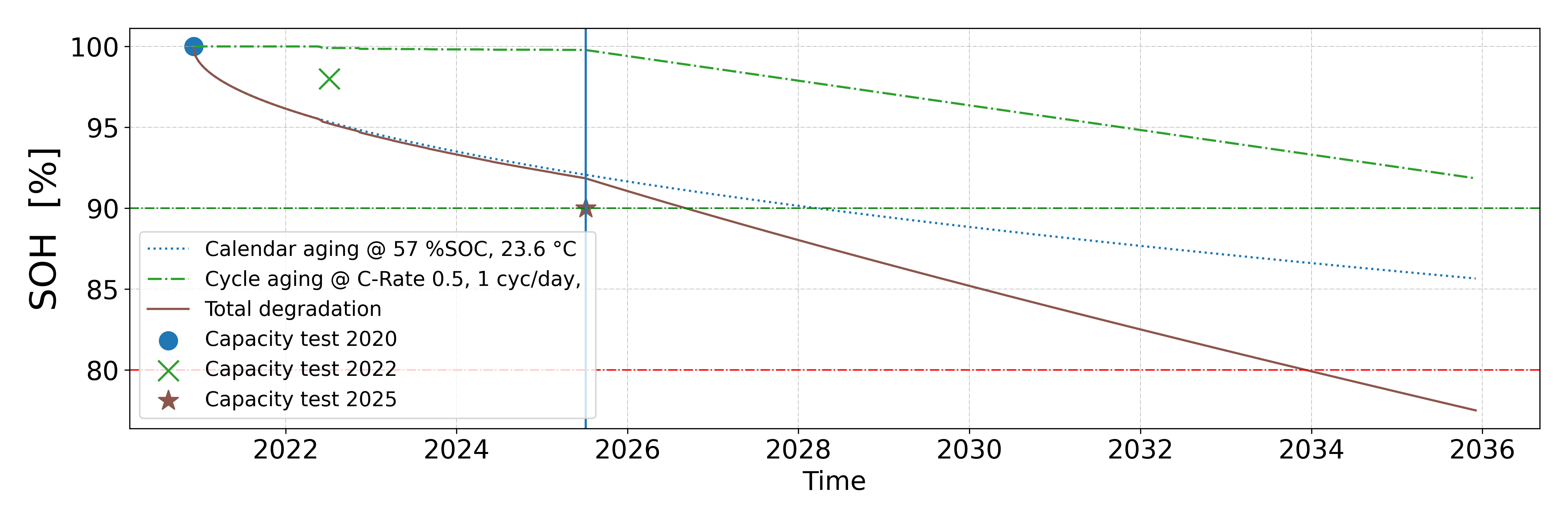}
    \caption{BESS capacity test results and SOH projection}
    \label{fig:degradation}
\end{figure}

The BESS is rated at \SI{55}{kW}/\SI{79}{kWh}, comprising eight packs of lithium-ion NMC cells, and ata are logged at a rate of 1 Hz and include active/reactive power, voltage, current, SOC, and temperature. As shown in Fig. \ref{fig:roundtrip}, multiple capacity test for the system is carried out on the BESS. With the AC power setpoint at 15 kW, the well-controlled behavior at high SOC but reduced behavior at low SOC is evident, where the reduced DC power can limit the power conversion system.
Based on extensive testing events, the market-usable performance of the BESS is summarized:
\begin{itemize}
    \item \textbf{Usable energy window:} the pack can reliably follow power setpoints only within SOC 12 \% – 95 \%; this corresponds to a market-usable energy of approximately \SI{52}{kWh} at the \SI{30}{kW} limit, which is reduced from converter specification \SI{55}{kW} by connection limitation.
    \item \textbf{Round-trip energy efficiency:} measured 98\% at 50\% of power (\SI{15}{kW}) and about 90\% at full power (\SI{30}{kW}).
    \item \textbf{Setpoint tracking accuracy:} 3 \% during charging and 7.6 \% during discharging from the external measurement.
    \item \textbf{Thermal limits:} average operating temperature \SI{23.6}{\celsius} (6–49 °C range) under forced-air cooling.
    \item \textbf{State of health:} With power of (\SI{15}{kW}), useful energy capacity decreased from \SI{72}{kWh} (2020) to \SI{65}{kWh} (2025), i.e. 90 \% SOH.
\end{itemize}
Experimental testing was retried between 2021 and 2025, and a simplified cycle-calendar battery SOH estimation is presented in Fig. \ref{fig:degradation}. Previously, the tested BESS were not used frequently, where the major degradation factor is calendar aging; however, a more extensive cycling usage is projected to dominate the cycle life. The empirical battery aging simulation is detailed in \cite{Zhao2022Data-drivenApplications}.
A combined calendar- and cycling-aging model tuned to these measurements indicates that the system can deliver roughly 250 MWh of remaining throughput before reaching 80 \% SOH, equivalent to about 1700 full cycles at \SI{30}{kW}.
These empirically derived parameters are directly used in the optimization framework to ensure physically meaningful scheduling.

\begin{figure}[t]
\centering
\subfloat[Upward setpoint change: 86\% in 18 s and 100\% in 21 s]{%
    \includegraphics[width=0.95\linewidth, trim=10 5 0 0, clip]{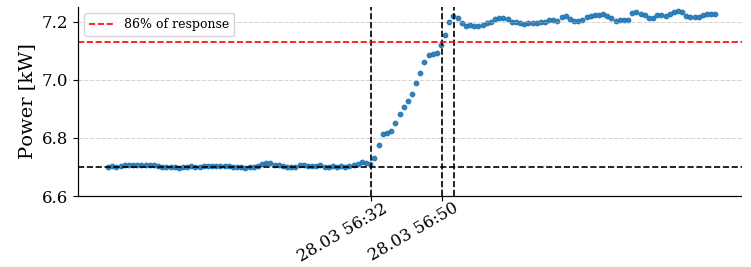}%
    \label{fig:elx_up}
}
\vspace{5pt}
\subfloat[Downward setpoint change: 86\% in 1.5 s and 100\% in 1.9 s]{%
    \includegraphics[width=0.95\linewidth, trim=0 2.5 0 0, clip]{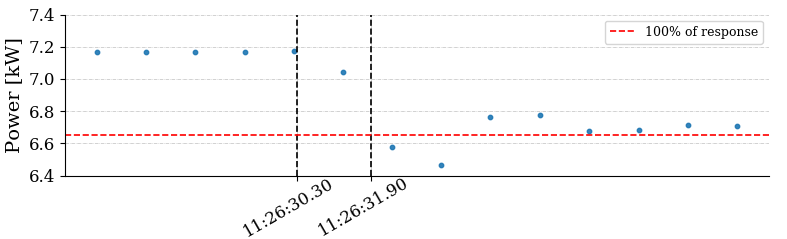}%
    \label{fig:elx_down}
}
\caption{Tested electrolyzer response behavoir}
\label{fig:elx_step_response}
\end{figure}

The electrolyzer installation consists of three modules, each rated at \SI{2.4}{kW} nominal power and designed for operation up to 30 bar. The system includes a dryer and a deionized water tank. Key experimental characteristics relevant for dispatch modeling are:
\begin{itemize}
    \item \textbf{Operating range:} 60–100 \% of nominal power; below 60 \%, the hydrogen–oxygen separation deteriorates, so the units must shut down.
    \item \textbf{Preheat phase:} electrolyte heating from ambient to \SI{45}{\celsius} follows an exponential process; this requires roughly 30 minutes with \SI{0.2}{kW} power without producing hydrogen.
    \item \textbf{Ramp dynamics:} ramp-up to full load in 21 s; ramp-down in 1.6 s, making the units suitable for up-regulation services in the power system, as shown in Fig. \ref{fig:elx_step_response}
\end{itemize}
Due to the rapid downward but slower upward response, the electrolyzer primarily qualifies for FFR and FCRN upregulation, resulting in reduced power consumption. Preheat requirements and power-range constraints are explicitly represented in the optimization model as binary‐state transitions between idle, standby, and run modes.

\section{Multi-Market Paticipation and Modeling}\label{sec:modeling}
The BESS operation is formulated as a mixed-integer linear program (MILP) that maximizes daily revenue from energy arbitrage and reserve-capacity provision. Decision variables include charging and discharging power, reserve capacity bids, and the SOE trajectory. The objective function is
\begin{equation}
\begin{aligned}
\max \sum_t \Big[
&(\lambda^{\mathrm{DA}}_t - \tau_{\mathrm{dis}})\, p^{\mathrm{dis}}_t
- (\lambda^{\mathrm{DA}}_t + \tau_{\mathrm{ch}})\, p^{\mathrm{ch}}_t  \\
&\quad + \sum_m \lambda^m_t\, p^m_t
- c_{\mathrm{deg}}\, E^{\mathrm{cyc}}_t
\Big]
\end{aligned}
\end{equation}
where $\lambda^{\mathrm{DA}}_t$ is the day-ahead energy price, $\tau$ the
charging/discharging tariffs, $p^m_t$ the reserve-capacity bids for market $m$, $c_{\mathrm{deg}}$ the marginal degradation cost, and $E^{\mathrm{cyc}}_t$ the equivalent full cycle energy used.
The battery SOE is calculated as,
\begin{equation}
\mathrm{SOE}_t
= \mathrm{SOE}_{t-1}
+ \eta\, E^{\mathrm{ch}}_t
- \frac{1}{\eta}\, E^{\mathrm{dis}}_t
\end{equation}
with round-trip efficiency $\eta$. Mutually exclusive charge/discharge operation
is enforced alongside SOE bounds (12--95\,\%), power limits, and a daily
throughput constraint that mitigates excessive degradation and unrealistic
cycling.

The electrolyzer scheduling problem is also cast as an MILP but includes unit
state transitions (idle, preheat, run) and nonlinear performance
characteristics. The objective maximizes hydrogen revenue net of electricity
costs, start/stop penalties, and income from reserve-capacity markets:
\begin{equation}
\begin{aligned}
\max \sum_{n,t} \Big[
&(\lambda^{\mathrm{H_2}} - \kappa_{\mathrm{run}})\, f_{t,n}
- \lambda^{\mathrm{spot}}_t\, p_{t,n} \\
&\qquad
+ \sum_{m}
\lambda^m_t\, p^m_{t,n}
- \kappa_{\mathrm{start,stop}}\, y_{t,n}\Big]
\end{aligned}
\end{equation}
where $m \in \mathcal{M}_{\mathrm{cap}}$ with
$\mathcal{M}_{\mathrm{cap}}=\{\mathrm{FFR},\, \mathrm{FCRDU},\, \mathrm{FCRN}\}$
denoting the available reserve-capacity markets, $\lambda^m_t$ the corresponding
capacity prices, and $p^m_{t,n}$ the reserve-capacity bid of module $n$. $\kappa_{\mathrm{start,stop}}$ represents the start-stop cost each time.

Both optimization problems are implemented in Python and solved with Gurobi. 
Historical price data for 2022–2025 were obtained from Energinet’s API, and activation factors were synthesized from frequency recordings following the procedure in \cite{Thingvad2022EconomicMarkets}. 
Hydrogen production is mapped from electrical power using the fitted quadratic flow model
$f_{t,n}=\alpha_2 p_{t,n}^2 + \alpha_1 p_{t,n} + \alpha_0$, which is valid only
when the module is in the running state. The preheat temperature follows the measured exponential law
$T(t)=T_0 + (T_\infty - T_0)\!\left(1 - e^{-k t}\right)$ with
$k=0.019~\mathrm{min^{-1}}$, enforcing a minimum warm-up duration of
approximately 30\,min before hydrogen production is allowed. Reserve-capacity
bids are constrained by the available headroom between $P_{\min}$ and $P_{\max}$
and by the required direction of response (up- or down-regulation). Additional logical constraints enforce state exclusivity, minimum up/down durations, and consistent start/stop identification.
For each day, a 24-hour horizon for BESS or a 96 × 15-minute horizon for electrolyzer is optimized, assuming perfect price foresight. While this yields an upper bound on achievable revenue, it enables the relative value of cycling, reserve participation, and physical constraints to be systematically compared.

\section{Results}\label{sec:results}

\begin{figure}[tb]
    \centering
    \includegraphics[width=\linewidth, trim=0 0 0 10, clip]{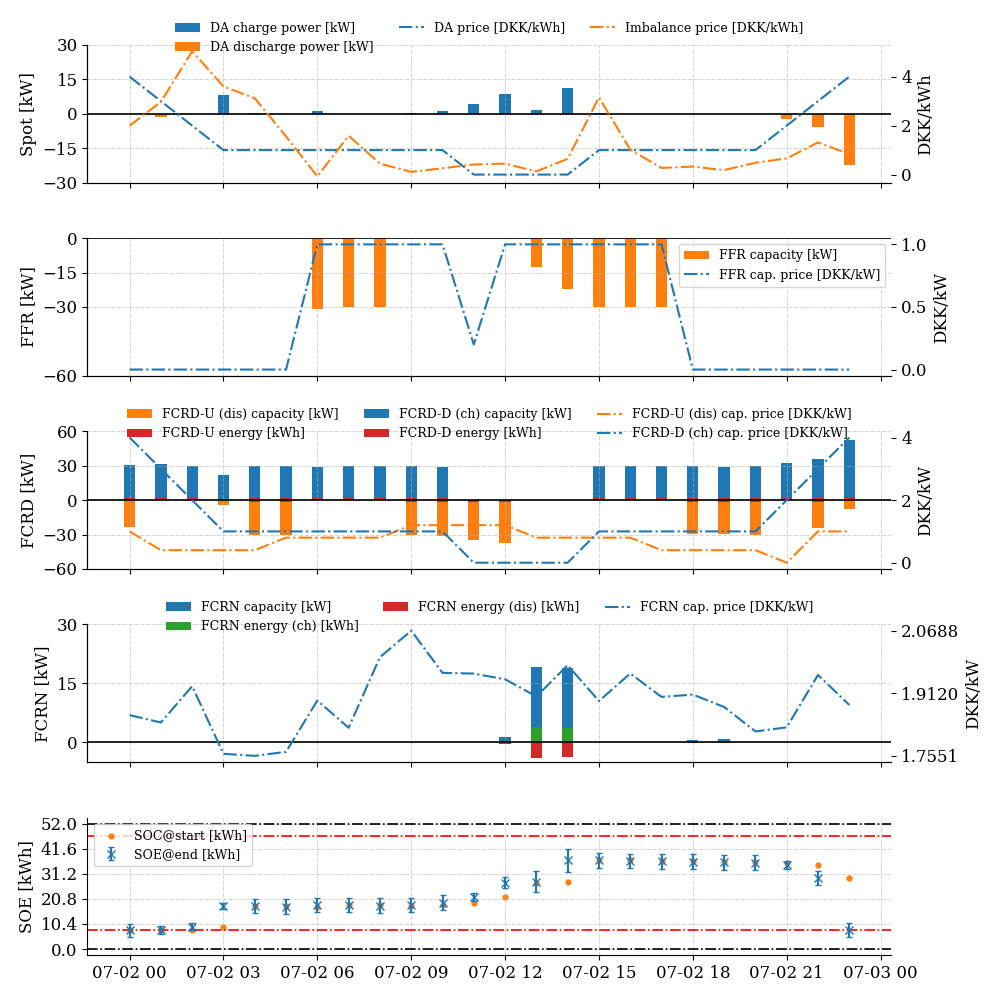}
    {\footnotesize \textit{*X-axis shows the time for 24h with 1 hour resolution (24 data points)}}
    \caption{Optimal BESS schedule in all available markets.}
    \label{fig:spot_ffr_fcrd_fcrn}
\end{figure}

The BESS is evaluated under daily throughput limits ranging from 0.1 to
5 cycles per day using three years of historical price data. For a nominal
limit of 1~cycle per day, the accumulated revenue reaches approximately
$6.2\times10^{5}$~DKK. Increasing the limit to 3--5~cycles per day provides
less than a 6\,\% revenue improvement, while accelerating degradation from
11\,\% to nearly 20\,\% SOH loss over the same period. These results indicate
that, in a multi-market environment dominated by reserve-capacity payments,
profitability is largely insensitive to additional cycling once sufficient
energy is allocated to maintain eligibility for ancillary services.
As shown in Fig. \ref{fig:spot_ffr_fcrd_fcrn} and TABLE \ref{tab:spot_ffr_fcrd_fcrn}, the operation schedule is detailed, where reserve activation is modeled assuming conversion of 20\% of FCRN bids and 5\% of FFR/FCRD bids into energy. Specifically, the spot market charges during the low DA price and sales at the end of the day. FFR is activated in the middle of the day when the price is higher. FCRD is very active throughout the day, whereas FCRN is activated for only a few hours. Finally, the SOE generally ships energy from daytime to nighttime. Regarding financial behavior, the majority of the revenue comes from FCRD-D, which accounts for more than 60\% of the overall revenue. The main energy throughput is due to the spot market, taking around half of the charging and discharging energy.
Revenue-stacking analysis shows that the optimizer relies primarily on FCRD and
FCRN products, using the day-ahead market to position the SOE ahead of
high-value reserve hours. The inclusion of realistic operational constraints
(12--95\,\% SOC window, 0.9 round-trip efficiency, and setpoint accuracy limits)
substantially reduces idealized profits but yields schedules that are feasible
for real hardware testing and consistent with PowerLabDK performance
characteristics.

\begin{table}[b]
\centering
\caption{Revenue and energy exchange in
Fig.~\ref{fig:spot_ffr_fcrd_fcrn}.}
\label{tab:spot_ffr_fcrd_fcrn}
\setlength{\tabcolsep}{4pt}
\begin{tabular}{lrrrrrr}
\toprule
\textbf{} & \textbf{Spot} & \textbf{FFR} & \textbf{FCRD-U} & \textbf{FCRD-D} & \textbf{FCRN} & \textbf{Totals} \\
\midrule
\textbf{DKK} & 103.17 & 216.07 & 289.01 & 1061.76 & 79.26 & 1749.27 \\
\textbf{kWh In}       & 37.51  & --     & 0.00   & 30.92   & 8.17  & 76.60   \\
\textbf{kWh Out}      & 31.60  & 10.80  & 17.10  & 0.00    & 8.17  & 67.67   \\
\bottomrule
\end{tabular}
\end{table}

\begin{figure}[tb]
    \centering
    \includegraphics[width=\linewidth, trim=5 55 0 60, clip]{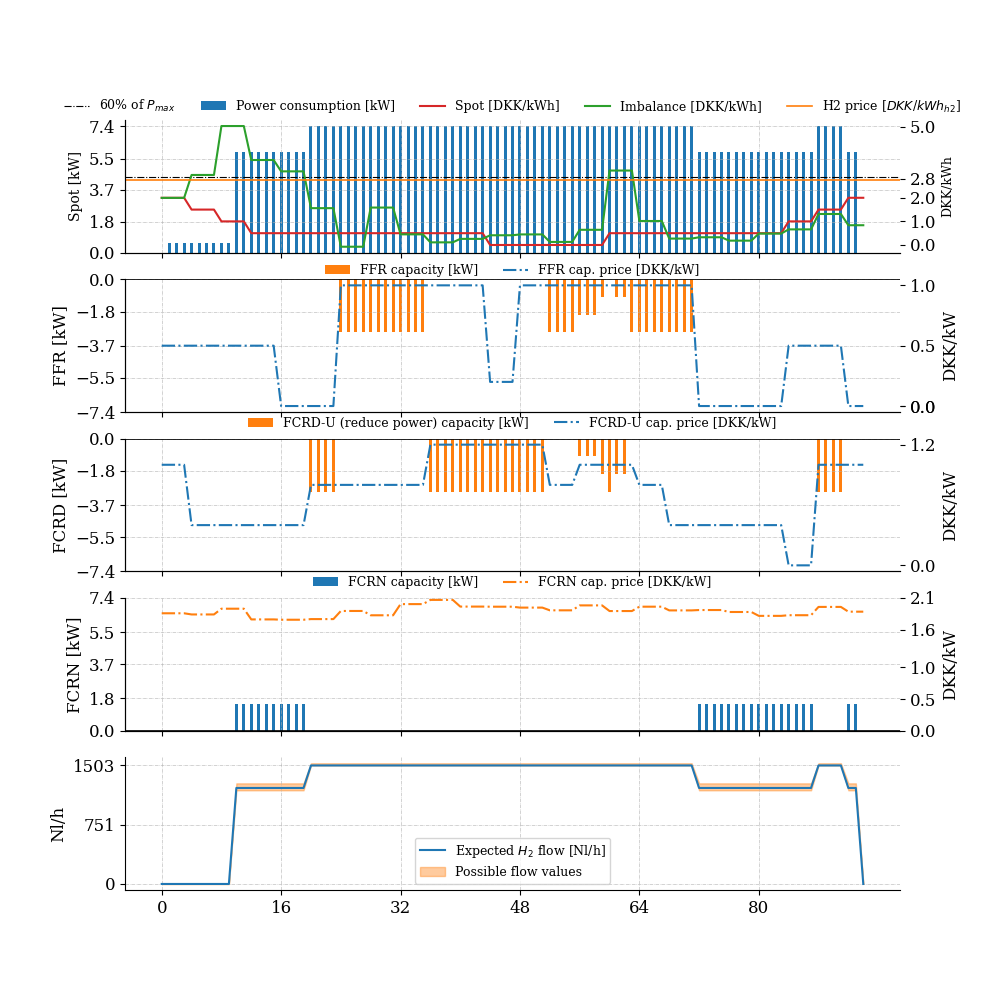}
    {\footnotesize  \textit{*X-axis shows the time for 24h with 15-minute resolution (96 data points)}}
    \caption{Optimal electrolyzer schedule in all available markets.}    \label{fig:electr_results_all_markets}
\end{figure}

For the electrolyzer case, the base-case simulations assume a hydrogen price of 6.61~€/kg (0.0044~DKK\, NL$^{-1}$). Under energy-only operation, pure hydrogen production yields only marginal profit due to the high electricity costs and non-negligible O\&M expenses. When ancillary service markets (FFR and FCRN) are included, the total daily profit increases by roughly 300\,\%, as the plant can monetize its fast down-ramping capability. The optimizer typically
schedules a 30-minute preheat during low-price hours, operates at full load during low or negative spot prices, and curtails consumption when reserve prices peak. The FCRN revenue includes energy payment (FCRN(E)) and capacity payment (FCRN(C)). For multi-market participation shown in TABLE \ref{tab:electr_results_all_markets}, the FFR, FCRD-U, and FCRN(C) are the most profitable items in the business case study.
As shown in Fig. \ref{fig:electr_results_all_markets}, Hydrogen flow is reduced during FCRN hours because the sale of reserve capacity is more profitable than producing hydrogen. The observed fluctuations in the flow curve reflect the stochastic nature of reserve activations.
Doubling the hydrogen price results in nearly a fourfold increase in total
revenue. The higher production hours expand the feasible reserve headroom,
allowing substantially larger FFR and FCRN bids. This illustrates the
nonlinear coupling between PTX revenues and flexibility-market participation.

\begin{table}[b]
\centering
\caption{Financial outcomes of in Fig. \ref{fig:electr_results_all_markets}.}
\label{tab:electr_results_all_markets}

\setlength{\tabcolsep}{3.5pt}
\begin{tabular}{lrrrrrr}
\toprule
\textbf{Item/[DKK]} & \textbf{H$_2$} & \textbf{FFR} & \textbf{FCRD-U}
& \textbf{FCRN(E)} & \textbf{FCRN(C)} & \textbf{Total} \\
\midrule
Revenue & 243.86 & 82.86 & 89.96 & 3.35 & 76.05 & 496.08 \\
Cost    & 204.95 & 0.00  & 0.00  & 1.67 & 0.00 & 206.62 \\
Profit  &  38.91 & 82.86 & 89.96 & 1.67 & 76.05 & 289.46 \\
\bottomrule
\end{tabular}
\end{table}

\begin{figure}[tb]
    \centering
    \includegraphics[width=1\linewidth]{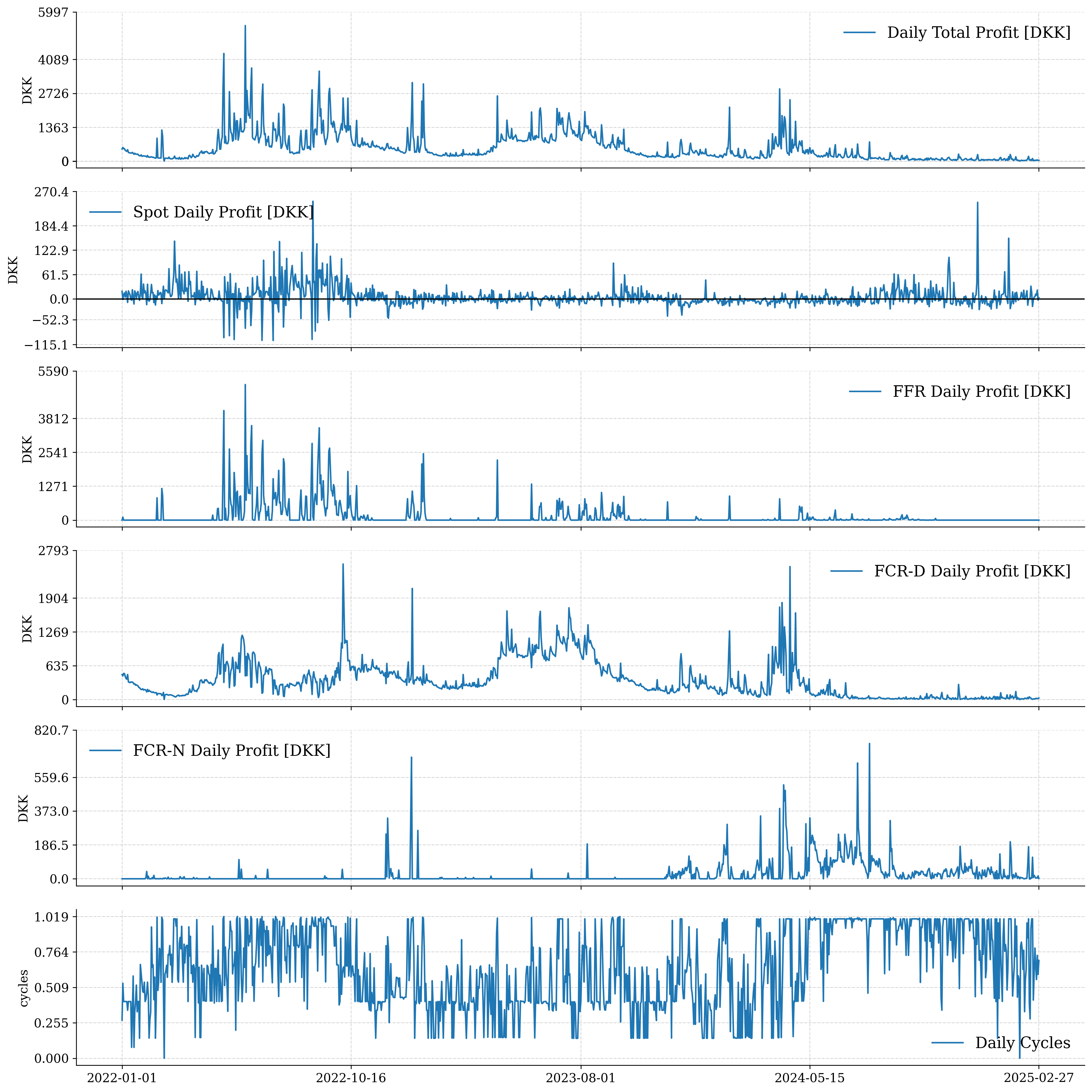}
    \caption{BESS case study with 1 cycle per day limit.}
    \label{fig:bat_basecase_finance_sim}
\end{figure}

\begin{figure}[tb]
    \centering
    \includegraphics[width=1\linewidth]{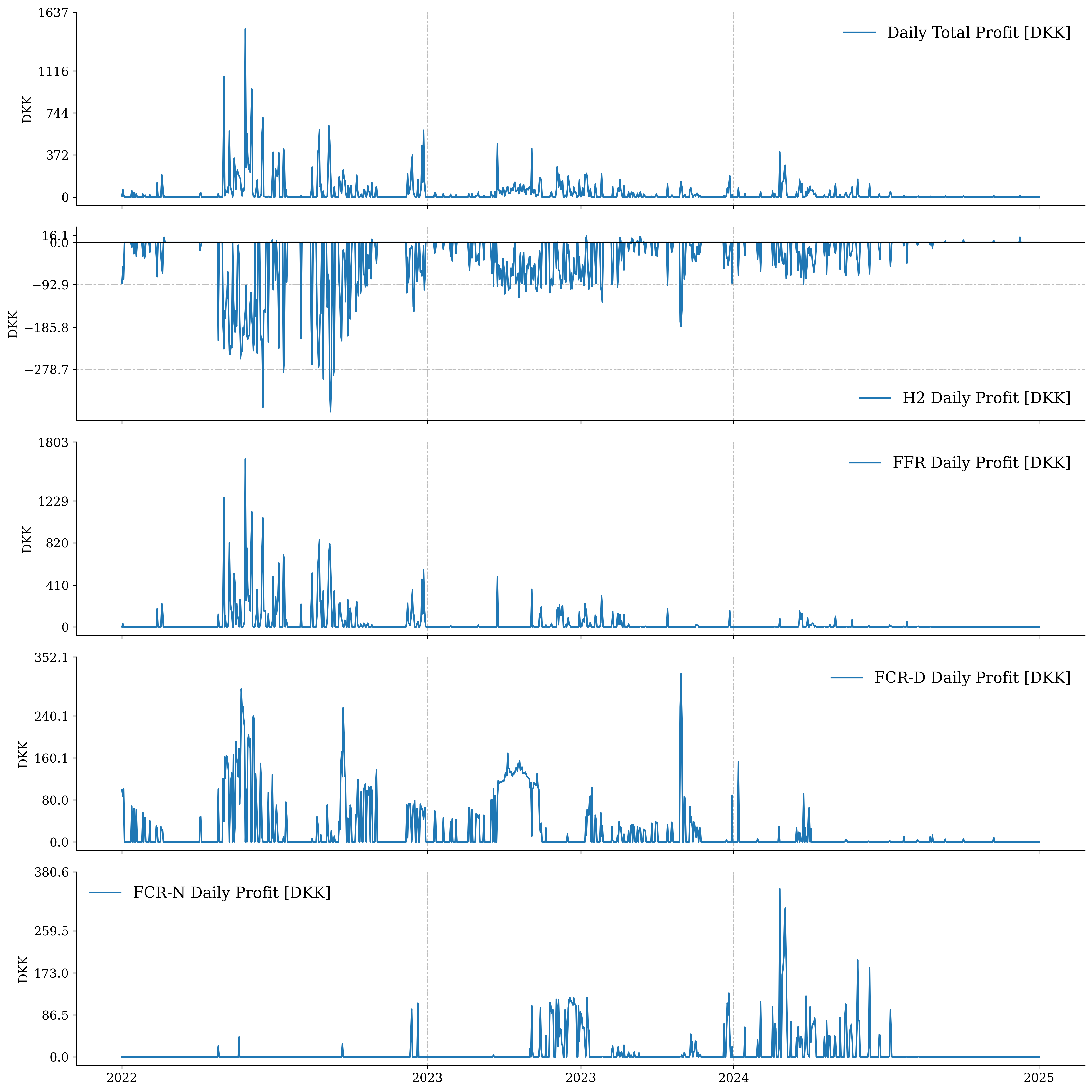}
    \caption{Electrolyser case study with 6.61 EUR/kg hydrogen price.}
    \label{fig:el_h2p_0044_h2c_0003_ss20}
\end{figure}

Moving forward from single-day optimization to multi-year simulation, our models run for each 24-hour period between 2022 and 2025 for BESS and electrolyser, as shown in Fig. \ref{fig:bat_basecase_finance_sim} and Fig. \ref{fig:el_h2p_0044_h2c_0003_ss20}. Significant variation is observed over time and across different revenue sectors. The daily revenue of BESS has generally been reducing over the years, although some peaks are imposed by the FCRD market contribution. The market dynamic is significant, where the FFR in 2022, FCRD in 2023, and FCRN in 2024 reach their peaks in turn.
It is interesting to observe that the production of pure hydrogen often leads to losses, as the price is quite low and barely covers the assumed operating costs, while at the same time, electricity prices for consumption are high. Nevertheless, the operation remains profitable, with profitability driven by the reserve markets, which follow similar trends to the BESS case study with market participation. 
\vspace{-1pt}
\section{Discussion}\label{sec:Discussion}
\begin{figure}[tb]
\centering
\subfloat[Monthly profits for the 1~cycle/day strategy]{%
    \includegraphics[width=\linewidth]{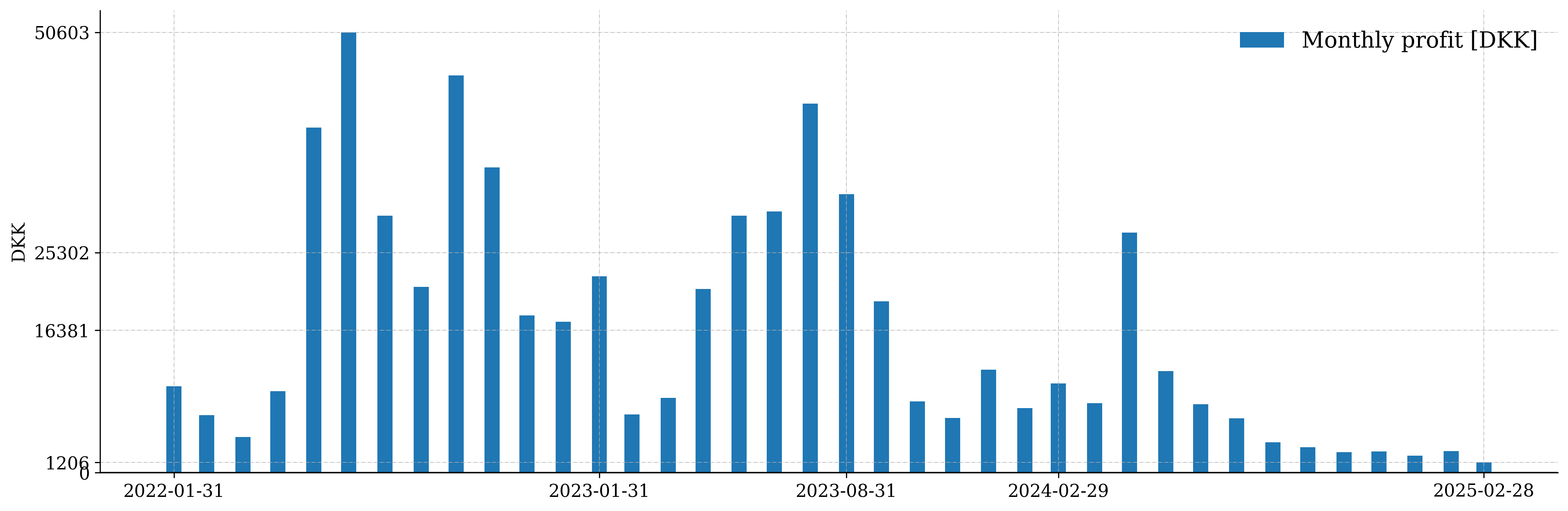}%
    \label{fig:1cpd_monthly}
}
\vspace{6pt}
\subfloat[Cumulative results of BESS with 1~cycle/day]{%
    \includegraphics[width=\linewidth]{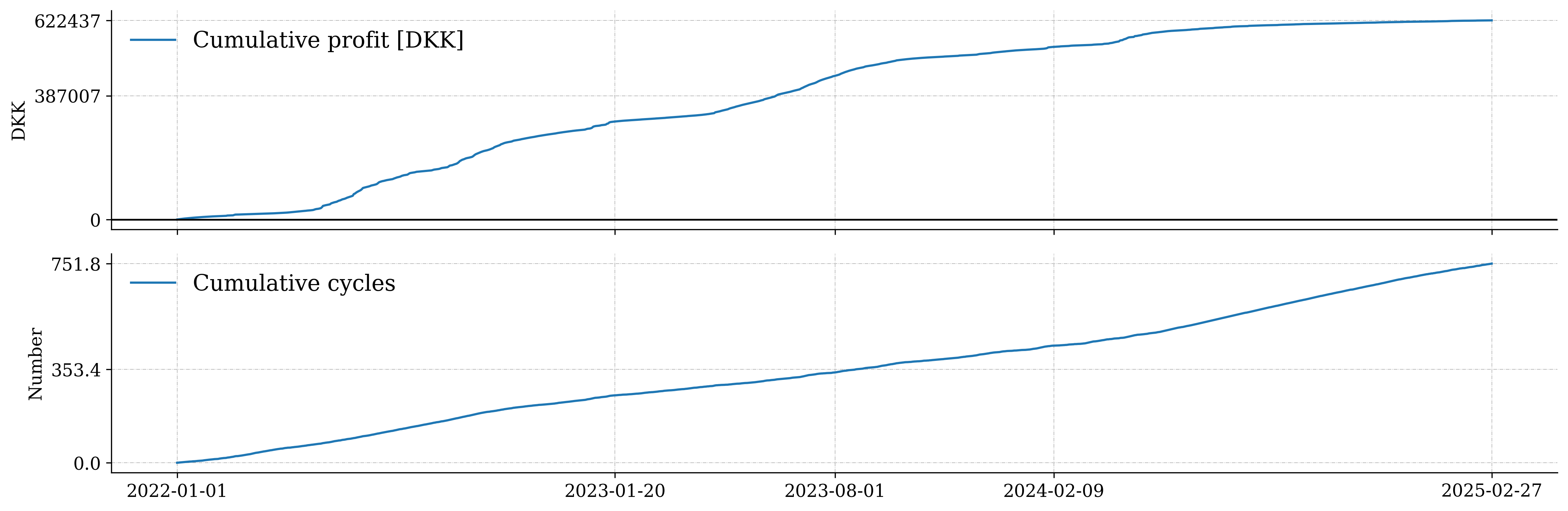}%
    \label{fig:cumulatives_1cpd_52kwh_mm_2021}
}

\vspace{6pt}
\subfloat[PTX business case comparison]{%
    \includegraphics[width=\linewidth]{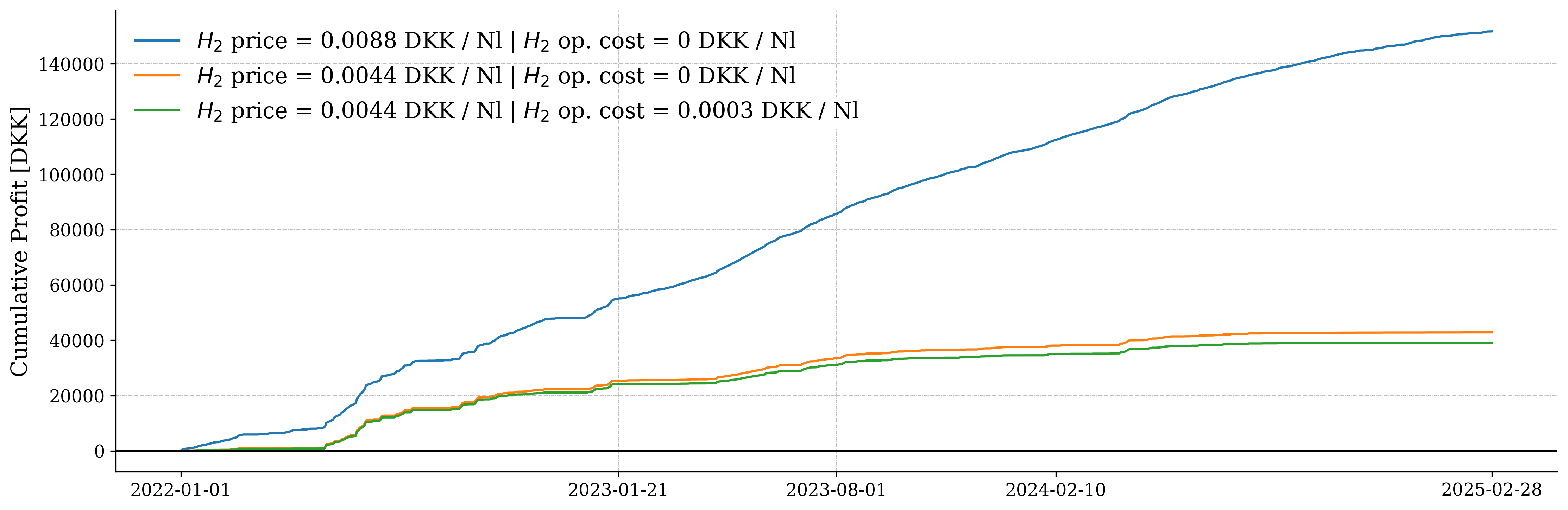}%
    \label{fig:ptx3year}
}
\vspace{-4pt}
\caption{Multi-year performance of the BESS and electrolyzer under multi-market participation.}
\vspace{-5pt}
\label{fig:combined_results}
\end{figure}

\vspace{-3pt}
The results of the monthly revenues are shown in Fig. \ref{fig:1cpd_monthly} with 1-cycle-per-day limit. It is evident that profits have been decreasing since mid-2023. Specifically, the spot-market results are often negative, where the battery uses this market to charge itself in preparation for reserve provision. BESS does not reach its daily cycling limit most of the time, as multi-market operations reserve most of the capacity, effectively limiting the flow of energy. The overall outcomes of this operation are given in Fig. \ref{fig:cumulatives_1cpd_52kwh_mm_2021}. The total profits amount to 622,437 DKK at the cost of 751 cycles and an 11\% decrease in the SOH within 3 years.
Furthermore, a sensitivity study is conducted on the electrolyzer case with varying hydrogen prices and operation costs, as shown in Fig. \ref{fig:ptx3year}, where an increase in hydrogen price significantly improves economics.
We also find neglecting setpoint accuracy or preheat delays can overestimate market revenues by 10 to 20\%.
The presented framework can be scaled up, and the combination of laboratory validation and market-based optimization provides a template for similar facilities and hybrid energy systems.

\section{Conclusion}\label{sec:conclusion}
This work presented an integrated experimental–modeling-optimization framework for evaluating multi-market operation of a \SI{55}{kW}/\SI{79}{kWh} BESS and a modular \SI{3}{\times}\SI{2.4}{kW} electrolyzer, embedding hardware performance such as the BESS’s \SI{52}{kWh} usable energy window, 90–98\% efficiency, and measured electrolyzer ramp times (21~s up, 1.9~s down). Results show that reserve markets dominate the value, and hydrogen production is only marginally profitable. However, participation in FFR, FCRD, and FCRN significantly increases daily profit. Overall, the study demonstrates that realistic modeling of physical performance and multi-market participation is crucial for assessing the economic potential of such assets.

\vspace{-1pt}
\section*{}

\end{document}